*Review*

# A Review of Knowledge Graph Completion

Mohamad Zamini [1,*], Hassan Reza [1] and Minou Rabiei [2]

[1] Department of Computer Science, University of North Dakota, Grand Forks, ND 58202, USA
[2] Department of Petroleum Engineering, University of North Dakota, Grand Forks, ND 58202, USA
* Correspondence: mohamad.zamini@und.edu

**Abstract:** Information extraction methods proved to be effective at triple extraction from structured or unstructured data. The organization of such triples in the form of (head entity, relation, tail entity) is called the construction of Knowledge Graphs (KGs). Most of the current knowledge graphs are incomplete. In order to use KGs in downstream tasks, it is desirable to predict missing links in KGs. Different approaches have been recently proposed for representation learning of KGs by embedding both entities and relations into a low-dimensional vector space aiming to predict unknown triples based on previously visited triples. According to how the triples will be treated independently or dependently, we divided the task of knowledge graph completion into conventional and graph neural network representation learning and we discuss them in more detail. In conventional approaches, each triple will be processed independently and in GNN-based approaches, triples also consider their local neighborhood.

## 1. Introduction

A graph can be directed if the order of nodes in the graph is important or undirected if the order of the nodes in the graph is not important. A knowledge graph is a heterogeneous multi-digraph which means it is directed, and multiple edges can exist between two nodes. An agent generates knowledge by relating elements of a graph to real-world objects and actions. A knowledge graph (KG), also known as a knowledge base, is a structured representation of facts that describes a collection of interlinked descriptions of entities, relationships, and semantic descriptions of entities. KGs, as a compelling abstraction, help organize structured knowledge by linking them from multiple sources. The difference between the knowledge base and knowledge graphs is the assumption of being less rigidly defined, structured, homogeneous, and stable schema breaks which empower knowledge graphs to be more scalable. The advantage of KG is the better representation of heterogeneous objects using a unified space to connect them.

Four crucial elements of machine learning models are learning, memory, knowledge representation, and reasoning. Although learning can be addressed by machine learning, current machine learning models require lots of data that, in many cases, cannot be provided by one person or group. In addition to lack of data suffering, current advances in artificial intelligence still cannot address 'data context'. Hence, a new approach to better create knowledge from data. After putting data into context, the lack of explainability in machine learning models is still an open challenge that needs to be better addressed. A knowledge graph is a promising approach that can better lead researchers to address these limitations.

The idea of structured knowledge in a graph was first introduced by [1] in 1988, and in 2012 this concept gained great attention after its usage in Google's search engine. Google's KG, one of the most important projects in knowledge graphs announced in 2016, holds over 70B facts [2]. The underlying technology is not publicly documented, but they used the schema.org standard and integrated Wikipedia, World Bank, Eurostat, etc. Cyc [3] is one of the early AI projects based on a knowledge base containing 1.5 M concepts in various areas, including healthcare, finance, and transportation. Cyc's knowledge base is represented by the CycL language and benefits micro theories for reasoning. OpenCyc released part of Cyc in RDF format until 2017. Facebook's entities graph with over 500 M facts was launched in 2010 and contains Facebook users' information including profile information, interests, and connections. The research on Semantic Web and linked data generated some open datasets, including the linked open data cloud, which are now primarily used as KG datasets. These datasets are mainly generated from Wikipedia using its massive amount of factual knowledge. DBPedia [4] is a de facto central dataset on the Semantic Web, which was first introduced in 2007 by extracting knowledge from Wikipedia pages with over 13 B RDF triples. FreeBase [5] was launched by a company in 2008, and Google bought it in 2010. Google shut down freebase in 2016, and its latest dump has over 1.9 B facts. In 2008 YAGO [6] also was first released. YAGO combines entities from Wikipedia articles with WordNet synsets. YAGO contains over 120 M facts. Wikidata [7] is another dataset launched in 2012 based on Wikipedia knowledge and revised by community members. Wikidata is structured by a customized data model supporting RDF and OWL with over 7B triples.

Entities can be real-world objects, events, and abstract concepts corresponding to a node, and directed edges are considered relationships. The knowledge graph stores objective information structured in RDF-style triples, which consists of two entities and one relation in the form of (*head, relation, tail*) or (*subject, predicate, object*) [8]. In a knowledge graph, labels are types of relations that can connect the facts; edges (relations) are specific facts connecting two nodes (head and tail entities).

Current real-world knowledge graphs are usually incomplete and need an inference engine to predict links and complete the missing facts among entities available in the KG. Relation classification or inference from information already available KG is called link prediction. The process of completing incomplete triples (i.e., (*Einstein, ?, Germany*)) is called knowledge graph completion (KGC). An example of it can be seen in Figure 1.

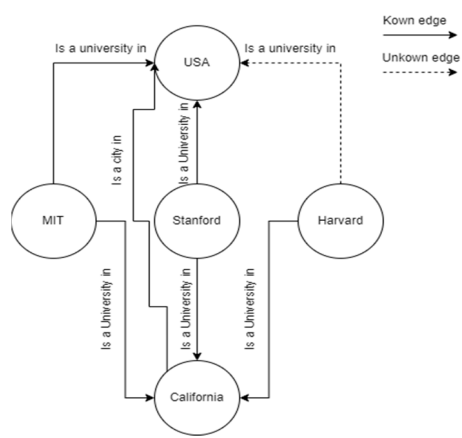

**Figure 1.** Sample KG where there exists a missing edge between the two nodes.

A common approach in link prediction and knowledge graph completion is via embedding into vector spaces to learn representations of entities and relations and embedding vectors of entities and relations can then be updated by maximizing the global plausibility. Embedding methods generalize from known facts and model triple-level uncertainty. Compared to traditional one-hot representation approaches, knowledge graph embedding can better address semantic computing using a distributed representation method. However, the reasoning results are not globally consistent. Different models have been proposed to solve this issue. Different scoring functions are defined to measure tri-ples' plausibility to enable updating the representation on the training data. Using different scoring functions in knowledge graph embeddings will reflect different designing criteria. We define representation models in the conventional knowledge graph completion section in Section 2.

Although representative models including TransE [9], TransH [10], DistMult [11], ComplEx [12], RotatE [13], show promise, they are incapable of leveraging the belief propagation of graph convolution in the representation learning process [14] and fail to cover the complex inherently implicit information in the local neighborhood of a triple. Instead, they treat each triple independently. As a result, they cannot benefit from graph structures to smoothen the embedding spaces. R-GCN [15], CompGCN [16], TransGCN [17], KE-GCN [14], and KBGAT [18] are some of the few works which leverage graph neural networks to jointly learn multi-layer layer representations of entities and relations. These works will be discussed in Section 3. Unlike most previous works which only cover conventional knowledge completion approaches, we covered graph neural network which has been recently studied in knowledge graph tasks. The challenges and gaps in Section 4 will be described, and in Section 5, the conclusion is expressed.

## 2. Conventional Knowledge Graph Completion

With unprecedented data volume growth worldwide, utilizing traditional graph structures to construct and manipulate KG is hard. The traditional formal logic reasoning is not tractable or robust in large-scale KGs. Several link prediction embedding models have been recently proposed to calculate semantic relations between entities in KG, which will be discussed in this section.

Knowledge representation learning (KRL) or knowledge graph embedding (KGE) is trying to map entities and relationships into a continuous vector space to capture better the semantic relation between entities in low dimensional space. For example, each entity $h$ in a KG can be represented by a point $h$ in vector space, and each relation $r$ can be modeled as an operation like projection, translation, etc., in the space. The embedding procedure in a given KG starts with randomly representing entities and relations in a vector space. With the help of an evaluation function, the plausibility of each triple will be evaluated at each iteration. Then, the embedding vectors of entities and relations update through optimization algorithms to maximize the global plausibility of facts.

In Table 1, we define the Notation and Problem Definition for each triplet fact ($h$, $r$, $t$) for the following sections.

**Table 1.** Notation and Problem Definition.

| Notation | Description |
| --- | --- |
| d | Vector |
| $W_r$ | The normal vector of hyperplane |
| r | Embedding vector of relation |
| h, t | Embedding vectors of head and tail |
| M | Projection matrix |
| ⟨□⟩ | Diagonal matrices |
| d | The dimensionality of an entity in embedding space |
| k | The dimensionality of relation in embedding space |
| Re | The real part of a complex value |
| ⊗ | Hamilton product |

Although various KGEs have been widely studied, most of them can only train an embedding model based on its observed triples. Hence, most current studies have focused on generalizing KGE models. KGEs for relation prediction can be classified into translational, decompositional, CNN-based, and graph neural network-based models [18]. In the following section, we will discuss each model separately.

*2.1. Translational Models*

Translational models interpret relations as simple translations over hidden entity representations. Translational distance models measure the plausibility of a fact and exploit distance-based scoring functions. The translational-based models try to find a low-dimensional vector representation of entities in relation to the translation of entities.

TransE [9] is one of the common translational models where both entities and relations are considered vectors in the same space. This model aims to model the inversion and composition patterns. Despite the simplicity of TransE, it cannot perform well in one-to-many, many-to-one, and many-to-many relations [10,19]. Although some of the complex models handle this issue, they are still not efficient in the process. For example, relation *Writer of*, might learn similar vector representations for *Harry Potter*, *Fantastic Beasts and Where to Find Them*, and *The Ickabog* which are all books by J. K. Rowling. However, these entities are different. To overcome this issue, extensions of TransE including TransH [10], TransR [19], TransD [20], TransM [21], and TransW [22] have been recently proposed which have different relation embeddings and scoring functions.

TransH [10] models a relation as a translating operation on a hyperplane with almost the same complexity as TransE. In this model, each relation is represented by two vectors, the norm vector of the hyperplane and the translation vector on the hyperplane. They addressed the issue of N-to-1, 1-to-N, and N-to-N relations by enabling each entity to have distinct distributed representations. The experiments on link prediction, triplet classification, and fact extraction on benchmark datasets like WordNet and Freebase show improvements compared to TransE.

TransE and TransH simply put entities and relations within the same semantic vector space. However, an entity may have multiple aspects and relations. Each relation might focus on a particular part of that entity that might be far from others. Additionally, entities and relations are entirely different objects, making them unsuitable to be represented in the same vector space. TransR builds entity and relations embeddings in separate vector space and then create a translation in the corresponding relation space. The comparison between TransR and the two previously introduced models shows significant improvements, including link prediction, triple classification, and relational fact extraction. TransR uses a projection matrix that projects entities from entity to relation space [19].

TransD uses two vectors to represent an entity or a vector. One represents the meaning of the entity or relation, and the other is used to construct a dynamically mapping matrix. This way, it covers both diversity of relations and entities. TransD is proposed to simplify TransR by eliminating matrix-vector multiplication operations and has fewer parameters, resulting in more applicability to a large scale. The evaluation of the model in link prediction and triplet classification outperforms the previously mentioned models. In TransD, each entity-relation pair has a unique mapping matrix. The elimination of matrix-vector operations in this model improved the performance.

TransM [21] leveraged the knowledge graph's structure. This model's optimal function deals with each triplet based on its weight. In this model, the transition model for triplets will be held the same as TransE, but the optimal function they proposed uses pre-calculated weight corresponding to the relationship. The main difference between TransE and TransM is it is more flexible when dealing with heterogeneous mapping properties of KGs by minimizing margin-based hinge loss function. The proposed model outperformed in link prediction and triplet classification tasks.

Recently, TransW [22] proposed using word embeddings for knowledge graphs embeddings to better deal with unseen entities or relations. Unlike previous works, which ignore the detail of the words within triples, TransW aims to enrich a KG by missing entities and relations using word embeddings. The linear combination of word embedding of entities and relations in this model leads to detecting unknown facts. The word embedding for relations and entities is calculated separately using the Hadamard product. The results outperformed compared to previous translational approaches.

RotatE [13] is another translational-based approach for KG representation learning. This model can infer different relation patterns of symmetry and antisymmetry. RotatE model defines each relation as a rotation from the source entity to the target entity in the complex vector space. Some relations are symmetric like marriage, and some are antisymmetric like filiation; some relations are inverse like hypernym and hyponym, and finally, some are composed of others like my dad's wife is my mom. How to infer these characteristics in KGs is essential to predicting missing links. Unlike the models mentioned above, RotatE aims to model and infer these characteristics at the same time.

HAKE [23] is a translational distance model with some similarities to RotatE [13]. Despite RotatE, HAKE aims to model the semantic hierarchy rather than modeling relation patterns. Unlike RotatE, which models relations as rotations that lead two entities to the same modulus, HAKE explicitly models modulus information which considers the depth of the tree as moduli and the distance function just considers the modulo part. In Table 2, the scoring functions of the introduced models have been described.

**Table 2.** Scoring functions of state-of-the-art translational-based knowledge graph embedding models.

| Model | Score Function | Memory Complexity |
| --- | --- | --- |
| TransE | $\|\|h + r - t\|\|_{l_1/l_2}$ | $O(N_e d + N_r d)$ |
| TransH | $\|\|(h - w_r^T h w_r) + d_r - (t - w_r^T t w_r)\|\|_2^2$ | $O(N_e d + N_r d)$ |
| TransR | $\|\|M_r h + M_r t\|\|_2^2$ | $O(N_e d + N_r(d^2 + d))$ |
| TransD | $\|\|(r_p h_p^T + I)h + r - (r_p r_p^T + I)t\|\|_2^2$ | $O(2N_e d + 2N_r d)$ |
| TransM | $w_r \|\|h + r - t\|\|_{l_1/l_2}$ | $O(N_e d + N_r k)$ |
| TransW | $\|\|(\sum h_i \otimes w_{hi} + b_h) + \sum r_i \otimes w_{ri} - (\sum r_i \otimes w_{ti} + b_t)\|\|_{1/2}^2$ | |
| RotatE | $-\|\|h \odot r - t\|\|$ | $O(2N_e d + 2N_r d)$ |
| HAKE | $\|\|h_m \circ r_m - t_m\|\|_2 + \lambda \|\|\sin((h_p + r_p - t_p)/2)\|\|_1$ | $O(2N_e d + 2N_r d)$ |

Compared to other models discussed in the following section, translational models are arguably faster, easier to train, and have fewer parameters to fine-tune. However, as the operations of translational models are mainly addition or multiplication, results lack expressiveness. The expressivity of the models for graphs can be defined as the diversity of their graph representations. Below the order of discussed models have been depicted in Figure 2.

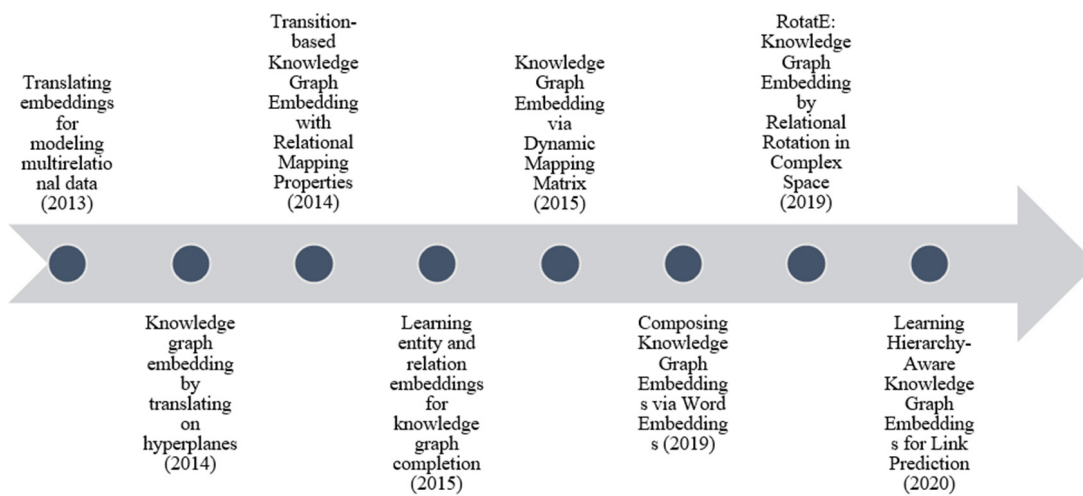

**Figure 2.** Translational model timeline.

*2.2. Tensor Dompositional Models*

The other types of embedding models are tensor decompositional models, which use tensor products to capture rich interactions. Tensor is a multi-dimensional numeric field that generates aliases scalars, vectors, and matrices [24].

Among tensor decompositional models RESCAL [25] and its extensions are more trendings. RESCAL follows a statistical relational learning approach to address uncertainty and complex relational structures. It uses a vector to capture to represent the latent semantics of each entity and a matrix that models pairwise interactions among latent factors [26]. The issue of this approach is the number of parameters which is $o(d^2)$. Additionally, RESCAL has never been tested on data with many relation types [27]. RESCAL has a large number of parameters, and this makes it prone to overfitting.

To simplify RESCAL, DistMult [11] proposed using bilinear diagonal matrices, which is a particular case of bilinear objective used in neural tensor models (NTN) and reduces the number of parameters to $o(d)$ per relation. NTN models are among the most expensive models as they use linear and bilinear relation operations. In RESCAL, each relation is represented by a square matrix, and DistMult simplifies it by using a diagonal matrix. The issue of DistMult is it can just deal with symmetric relations because of using diagonal matrices. Other than that, is a model, characterized by three-way interactions between embedding parameters to produce a single feature per parameter. The linear transformation on entity embedding vectors cannot model asymmetric relations. Using such models learn shallow features with less expressive features, but it is scalable to large knowledge graphs [28].

ComplEx [12] also aimed to generalize DistMult by proposing complex-valued embeddings to improve asymmetric relations modeling between entities. The dot product of vector embeddings cannot well represent asymmetric relations. In asymmetric relations, we cannot interchange subjects and objects without changing the relation. ComplEx embedding method infers new relational triplets with asymmetrical Hermitian product. In this model, entity and relation embeddings are in a complex space rather than one real space. This enables ComplEx model asymmetric relations.

Suchanek et. al. [29] subsumes DistMult and ComplEx with more generalizability by exploiting hypercomplex space for learning KG embeddings. Unlike standard vector space with single component *i*, each quaternion embedding is a vector in the hypercomplex space H with the imaginary components *i*, *j*, and *k* with a new scoring function with relational quaternion embedding through the Hamilton product. It has been proved that the Hamilton operator compared to Hermitian and inner product in Euclidean space, has better expressiveness. However, rotation-based models cannot model hierarchical structure. This model also is not capable of modeling multiple relations between two entities at the same time. A new method called dual quaternion KGE (DualE) [30] is proposed to solve the issues as mentioned above. Embeddings in dual quaternion space are vectors in hypercomplex space. This model integrates and unifies translation and rotation operations.

Tucker [31] employs a different decomposition model called a Tucker Decomposition to compute a smaller core tensor and a sequence of three matrices where each matrix represents entity embedding and relation embedding separately. In Table 3, the scoring functions of the introduced models have been described.

**Table 3.** Scoring functions of state-of-the-art tensor decompositional-based knowledge graph embedding models.

| Model | Score Function | Memory Complexity |
| --- | --- | --- |
| RESCAL | $h.W_r.t$ | $O(N_e d + N_r d^2)$ |
| DistMult | $\langle h, r, t \rangle$ | $O(N_e d + N_r d)$ |
| ComplEx | $Re(\langle h, r, \bar{t} \rangle)$ | $O(2N_e d + 2N_r d)$ |
| Quaternion | $h \otimes r^{\triangleleft}.t$ | $O(N_e d + N_r d)$ |
| DualE | $h \otimes r^{\triangleleft}.t$ | $O(N_e d + N_r d)$ |
| Tucker | $W \times_1 h^T \times_2 M_r \times_3 t$ | $O(N_e d + N_r d + d_e d_r d_e)$ |

The issue of the above-mentioned models is that the learned embeddings should be solely compatible with each individual fact. In another word, the above-mentioned models only consider structural information observed in triples rather than external information. This results in the incompatibility of downstream tasks [32]. This motivates researchers to include other information like entity types [33], logical rules [32], and relation paths [34] to improve the learning of embeddings. Below the order of discussed models have been depicted in Figure 3.

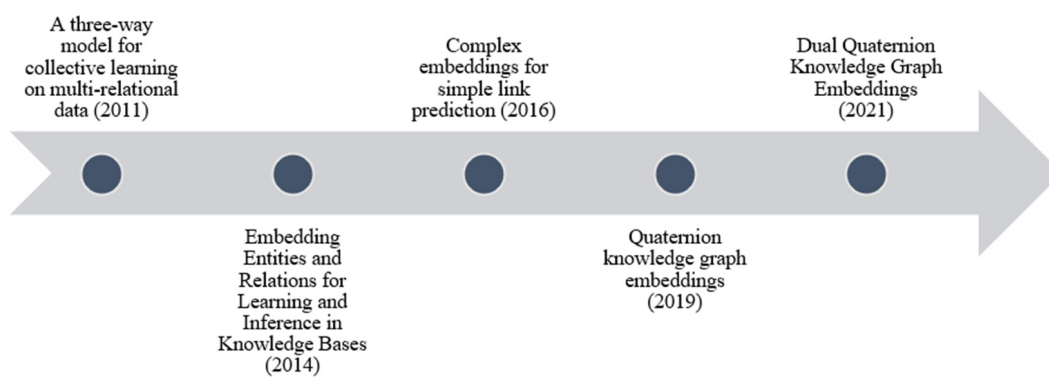

**Figure 3.** Tensor decompositional models' timeline.

### 2.3. Neural Network Models

Semantic matching energy (SME) [27] is one of the early neural network models which first project entities and relations to their vector embeddings in the input layer. The proposed model captures the inherent complexity in the data by defining similarities among entities and relations. The energy function is encoded using a neural network to extract relevant components of each argument's embedding using relation types. The computed results can be comparable in a space. It works based on energy function to assign low energies to plausible triplets of the multi-relational graph. Head entity and relation are combined and make a function; tail entity and relation also make another function in the hidden layer. Then they calculate the fact score by their dot product of two functions.

Neural Tensor Network (NTN) [35] is an expressive neural tensor network that is capable of reasoning over relations between entities. In this work, unlike previous works, they represent each entity as the average of its word vectors. They used a bilinear tensor layer that directly related two entity vectors across multiple dimensions. The model computes the relation by an NTN-based function. The scoring function of this model is:

$$f_r(h,t) = u_r^T \tanh(h^T M_r t + M_{r,1} h + M_{r,2} t + b_r) \quad (1)$$

The issue with these models is they learn more shallow and less expressive features than multi-layer models, which limits the performance of KGs. Link prediction should be manageable with the number of parameters and computational costs to be useful for KGs. They sacrifice accuracy over speed by using simple operations like inner products and matrix multiplications over an embedding space [28]. To increase the expressiveness, it is essential to increase the embedding size. However, increasing the embedding size is proportional to the number of entities and relations that exist in the graph. An intuitive example will be a model like DistMult with the embedding size of 200 on the Freebase dataset will require 33 GB of memory regarding its parameters[28]. Using fully connected models in multi-layer KGEs can be prone to overfitting. To solve that, convolutional layers evolved, which are highly optimized to GPU. In Table 4, the scoring functions of the introduced models have been described.

**Table 4.** Scoring functions of state-of-the-art tensor decompositional-based knowledge graph embedding models.

| Model | Score Function | Memory Complexity |
|---|---|---|
| SME | $g_{left}(h,r)^T g_{right}(r,t)$ | $O(N_e d + N_r d)$ |
| NTN | $r^T \tanh(h^T \hat{M} t + M_{r1} h + M_{r2} t + b_r)$ | $O(N_e d + N_r d^2 d)$ |

*2.4. Convolutional-Based Models*

Although previous models were fast models and can be scaled to large KGs, they learn less expressive features than multi-layer models. ConvE [28] yields the same performance as DistMult and R-GCN [15] with much fewer parameters and is effective at modeling nodes with high indegree. It uses 2D convolution layers for link prediction, which consists of a convolution layer, a projection layer that deals with embedding dimension, and an inner product layer. ConvE generates a matrix by wrapping each vector over several rows and concatenating the matrices. Each convolutional filters generate different feature map tensors to extract the global information. Although this model outperforms common convolutional models in computer vision and other areas is still shallow and needs to study deeper models to improve its performance. HypER [36] also applies convolutions but uses a fully connected layer to avoid wrapping and generate relation-specific convolutional filters.

The number of interactions that ConvE can capture between relation and entity embeddings is limited. To increase this number, InteractE [37] is proposed based on multiple permutations to capture possible interactions better, substituting simple feature reshaping used in ConvE with checked reshaping and circular convolution to capture more feature interactions in a depth-wise manner. This way the interaction between entity and relation embeddings for learning better representations and circular convolution will be enhanced.

ConvKB extends ConvE by omitting the reshaping operation in encoding representations in the convolution operation [38]. ConvKB [39] uses CNN to capture global relationships and transitional characteristics between entities and relationships. Each triple in this model is represented by a 3-column matrix where each column represents a vector of each element of a triple. The matrix is the input of a convolution layer to map to different feature spaces and then concatenate them to create a single feature vector as an input triple representation. Its plausibility score is calculated via the dot product of the feature vector and a weight vector. Despite ConvE, ConvKB covers global relationships between the same dimensional entries of an embedding triple.

ConEx [40] is a Hadamard product composition of a 2D convolution followed by an affine transformation and a Hermitian inner product on complex-valued embeddings. The proposed model uses the asymmetric properties of Hermitian products and the parameter sharing property of a 2D convolution. Hermitian product has been previously used in the ComplEx embedding model, which shows good expressiveness. However, results show that the Hamilton product is more expressive.

Although both previous convolutional models are parameter efficient and showed outperforming results, they consider each triple independently without considering the possible relationships between triples. Another challenge in embedding approaches mentioned above is the failure to capture multistep relationships; most of them solely work on the observed facts. They train each node and edge representation based on the context of triples they are involved in. An alternative approach is using machine learning architectures for graphs instead of computing numerical representations for graphs. Graph neural networks have been studied widely as a solution to these challenges. In graph neural networks, edges serve as weighted connections, and nodes serve as neurons. In Table 5, the scoring functions of the introduced models have been described.

**Table 5.** Scoring functions of state-of-the-art convolutional-based knowledge graph embedding models.

| Model | Score Function | Memory Complexity |
|---|---|---|
| ConvE | $f(vec\left(f([\bar{h}; \bar{r}]) \star \Omega\right))W)t$ | $O(N_e d + N_r d + Tm_\Omega n_\Omega + Td(2d_m - m_\Omega + 1)(d_n - n_\Omega + 1))$ |
| ConvKB | $concat(f([h, r, t]) \star \Omega))w$ | $O(N_e d + N_r k + 4T)$ |
| HypER | $f(vec(h \star vec^{-1}(w_r H))W)t$ | $O(N_e d + N_r d)$ |
| InteractE | $f(f(perm([h; r]) \circledast w)W + b)t$ | $O(N_e d + N_r d + Tm_\Omega n_\Omega + 2Tpd^2)$ |
| ConEx | $Re(\langle conv(h, r), h, r, \bar{t} \rangle)$ | |

The proposed models in this section can learn representations of entities and relations in KGs for link prediction. The effectiveness of these models depends on their ability to infer different relation patterns, including symmetry, asymmetry, composition, inversion, and transitivity. However, none of the current approaches can cover them well [41]. Recently, graph neural networks have been widely studied and seem to be a promising approach to solving this issue. In the next section, we discuss these approaches. Below the order of discussed models have been depicted in Figure 4.

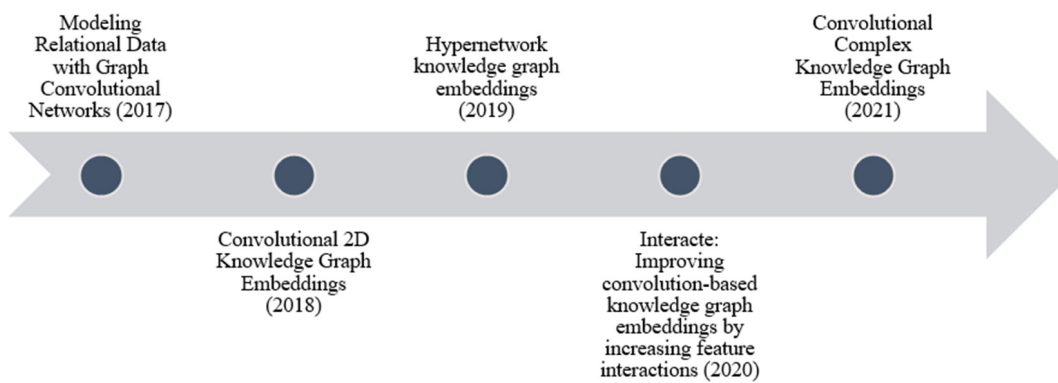

**Figure 4.** Convolutional based models' timeline.

## 3. Graph Neural Networks

Deep learning approaches have been exploited for graph data modeling and representation. An essential step in performing graph-structured data tasks is learning better representations. KGs can be treated as graphs with triplets in which relations are edges. Conventional neural networks are limited to handling only Euclidean data. By leveraging representation learning, graph neural networks have generalized deep learning models to perform on structural graph data with good performance. In GNNs, an iterative process propagates the entity state until equilibrium. This idea was extended by [42] to use gated recurrent units in the propagation step. Graph neural network (GNN) models have been proved to be a powerful family of networks that learns the representation of an entity by aggregation of the features of the entities and neighbors [43]. In traditional GNNs, multiple layers are stacked to aggregate information throughout the knowledge graph and output learned entity embeddings. Some GNN models also can learn relation embeddings.

GNNs have been recently applied in knowledge graphs to learn powerful embeddings by using topological structures in the KGs. GNNs generally update node representations by aggregating and propagating node features in the graph. Unlike conventional embeddings, GNNs are capable of end-to-end supervised learning, which can perform various classification tasks[44]. However, most of the commonly designed GNNs are suitable for uni-relational connections between entities which is not suitable for KGs which are multi-relational.

Several attempts have been exploited to apply neural networks to deal with structured graphs. Recursive neural networks were laid as early work to process data in acyclic graphs [45]. This idea has been extended to graph neural networks in [46] to generalize recursive graph neural networks for directed and undirected graphs. They generally learn the target node's representation by its neighbor's information iteratively until it reaches an equilibrium point. Graph neural networks with the key factor of high dimensional data growing have been widely studied and applied to learn representations from complex graph-structured data with remarkable performance in different domains.

Overall, graph-based models regarding their embedding dimensionality can be classified into low-dimensional and high-dimensional embedding. Low-dimensional embeddings of nodes in large graphs proved extremely useful in different prediction tasks. However, most existing approaches require all nodes in the graph to be present when training the embedding model. These approaches are transductive and are not capable of being generalized to unseen nodes. Another group of approaches has been recently studied, which are inherently inductive and generate node embeddings from previously unseen data. Generating new node embeddings in the inductive setting is more difficult because generalizing to unseen nodes requires aligning newly observed subgraphs to the node embeddings [47]. BoxE [48] and GraphSAGE [47] are examples of inductive embedding. GraphSAGE leverages node feature information to generate node embedding for unseen data in undirected graphs. It operates by sampling a fixed-size neighborhood of each node and applying an aggregator over it. However, GraphSAGE is incapable of distinguishing different graph structures. BoxE modeling is region based supervised embedding learning which embeds entities as points and relations as a set of hyper-rectangles to spatially characterize logical rules and more straightforward calculations in similarity representations. This model makes use of boxes in its loss function model, and the entities are still in the form of vectors. As a result, they cannot benefit from the probabilistic semantics of box embeddings. Below the order of discussed models have been depicted in Figure 5.

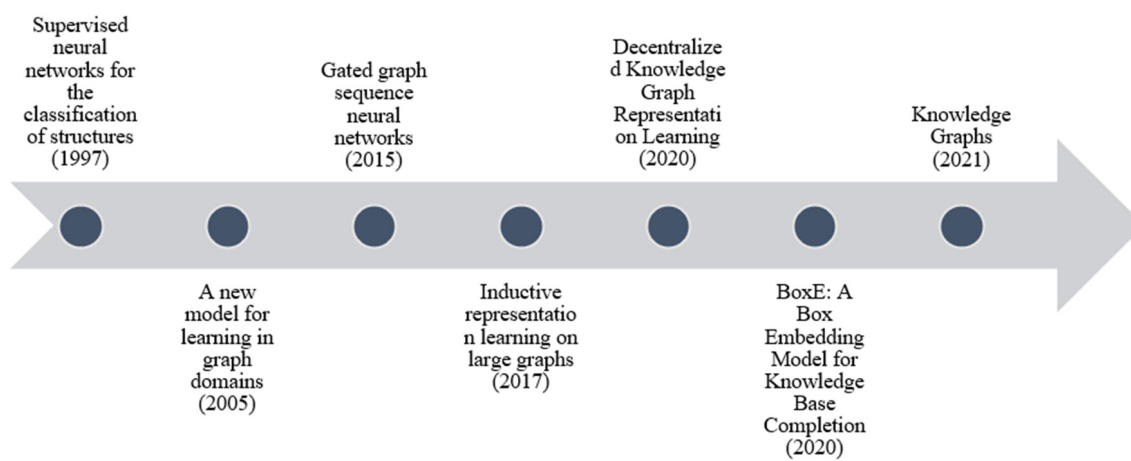

**Figure 5.** GNN-based models' timeline.

*3.1. Graph Convolution Network Models*

By evolving CNNs, convolution is exploited on graph data in parallel. Convolutional graph neural networks (ConvGNNs) are mainly divided into two main approaches: spectral based, which depends on graph structure based on the Laplacian eigenbasis, and spa-tial based, which works on sampling a fixed-size neighborhood of each node and aggregating over it. This approach has proved powerful in several large-scale inductive benchmarks [49].

Graph convolutional networks have gained much attention; they work under an encoder-decoder framework to aggregate local information in the graph neighborhood for each node. Similar to convolutional neural networks, which operate over local regions of input data, GCNs go over a node and its neighbors in the graph to learn the embedding representation of that node via recursive aggregation of embeddings of the neighborhood. Depending on the number of convolution layers, GCNs can capture information of immediate neighbors or K hops away nodes [50]. In other words, convolutions can smoothen the learned representations vectors of nodes over the entire graph.

In GCNs, the convolution operator uses locality information in graphs to leverage attributes associated with nodes [51]. However, general GCN is limited to undirected graphs and is not suitable for highly multi-relational data [28]. Additionally, it is limited to learning node embedding on fixed edges instead of jointly learning optimal embeddings of nodes and edges [14]. Relational Graph Convolutional Network (R-GCN) [15] is one of the well-known approaches which is developed for knowledge base completion tasks such as link prediction and entity classification. GCN represents a graph encoder, but to make it capable of specific tasks, it needs to be developed into R-GCN, which takes the neighborhood of each entity equally by hierarchical propagation rules to be suitable for directed graphs. It aggregates relation-specific transformation matrices with neighborhood information [52]. The encoder maps each entity to a real-valued vector, and the decoder (scoring function) reconstructs the edges of the graph based on vertex representations. In this model, DistMult factorization is used as a scoring function where every relation is related to a diagonal matrix. Optimizing cross-entropy loss pushes the model to pick better triples than negative ones. However, R-GCN does not take relation or attribute similarity between entities into account. Additionally, using R-GCN's scoring function generates many negative triples for a positive triple [53]. RA-GCN [54] has been proposed to solve these issues by improving the propagation extension for entity updating and extracting additional entity and related information through aggregation.

Unlike R-GCN which entity embedding learning was done through a convolution-based encoder and relation embedding learning was in the decoder, TransGCN [17] trains relation and entity embeddings simultaneously during graph convolution operation with fewer parameters compared to R-GCN by using relation as transformation operator on between head and tail entity in a triple. They used transE and RotatE on both datasets and RotatE-GCN showed better results than TransE-GCN. However, in TransGCN, the relation embedding during learning ignores entity representations. To solve this issue, KE-GCN [14] leveraged the strength of the GCN model and KGC methods for relation and entity embedding updates. KE-GCN is a heterogeneous learning model focusing on jointly propagating and updating knowledge embedding of both nodes and edges. Similarly, COMPGCN [16] also uses joint vector representation learning for nodes and edges in multi-relational graphs by leveraging various entity-relation composition operations from KGE models.

The entity and relation update of each model is summarized in Table 5. In this table, $h_i$ is hidden layer representations update entity $v_i$ in layer l. $W_r^{(l)}$ is the relation-specific weight matrix of layer *l*-th. $\sigma$ denotes a nonlinear activation function and $N_i^r$ denotes the set of neighbor indices of a node I with relation r. $W_r^{(l)}$ and $W_0^{(l)}$ also defined as $\sum_{b=1}^{B} a_{rb}^{(l)} V_b^{(l)}$ which is a linear combination of basis transformation $V_b^{(l)}$ with coefficients $a_{rb}^{(l)}$. $z_r^l$ which is initial relation representation can be defined the same as $W_0^{(l)}$ but instead of matrices, COMPGCN uses embedding vectors. As a result, $V_b^{(l)}$ is a set of learnable basis vectors and $a_{rb}^{(l)}$ is relation and basis-specific learnable scalar weight. *N* is also the set of immediate entity relation neighbors of entity v. $m_r^{l+1}$ is the aggregation representation of neighbors. In Table 6, the scoring functions of the introduced models have been described.

**Table 6.** Graph Neural Network-based update functions.

| Model | Relation Update | Entity Update |
|---|---|---|
| R-GCN | - | $h_i^{(l+1)} = \sigma(h_i^{(l)}W_0^{(l)} + \sum_{j \in N_i^r}\sum_{r \in R}\frac{1}{c_{i,r}}h_j^{(l)}W_r^{(l)})$ |
| RA-GCN | - | $h_i^{(l+1)} = \sigma(h_i^{(l)}W_0^{(l)} + \sum_{j \in N_i^r}\sum_{r \in R}h_j^{(l)}W_r^{(l)})$ |
| TransE-GCN | $r_k^{l+1} = \sigma(W_1^{(l)}r_k^{(l)})$ | $v_i^{(l+1)} = \sigma(\sum_{j \in N_i^r}\sum_{r \in R}\frac{1}{c_{i,r}}v_j^{(l)}W_r^{(l)} + W_o^{(l)}v_i^{(l)})$ |
| KE-GCN | $h_r^{l+1} = \sigma_{rel}()$ | $m_v^{l+1} = \sum_{(u,v) \in \mathcal{N}(r)} W_r^l \frac{\partial f_r(h_u^l, h_r^l, h_v^l)}{\partial h_r^l}$ |
| CompGCN | $h_r^{l+1} = h_r^l W_{rel}^l$ | $h_v^{l+1} = \sum_{(u,r) \in \mathcal{N}(v)} W_r^l \phi_{in}(h_u^l, h_r^l)$ |

Although the proposed GCNs can effectively improve accuracy, scalability is a major challenge for them. PinSage [55] can be a promising solution that is capable of handling billions of nodes and edges however, to the best of our knowledge, it has not been tested for knowledge graphs. In this approach, they select a fixed number of neighbors for all given nodes, which might result in dropping out some neighboring nodes and information loss in multi-relational data. Below the order of discussed models have been depicted in Figure 6.

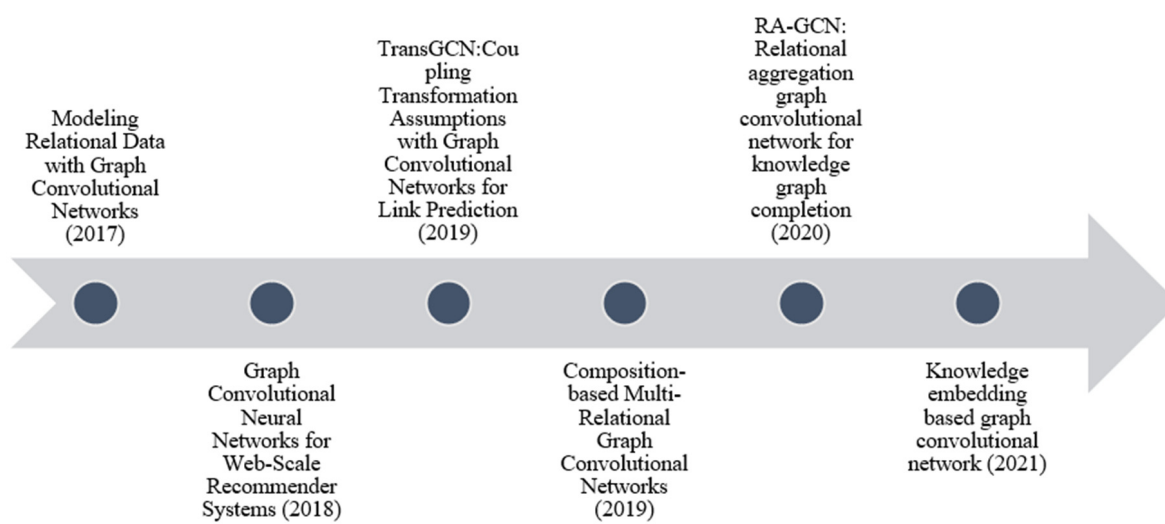

**Figure 6.** CNN-based models' timeline.

*3.2. Attention Neural Network Models*

Unlike the application of CNNs in images with a predictable number of neighbors, in graph data, the neighborhood of nodes in a graph is unpredictable. Hence, one of the challenges of the previous approaches is to define an operator which can be adaptive to the size of neighborhoods and hold the weight-sharing property of CNNs. New emerging research in this area is focused on more contextualized embeddings by exploiting information from their neighborhood. To address this challenge, an attention mechanism is proposed to learn those nodes that have more essential features than the current node. Attention mechanisms have been widely used in sequential tasks. Attention mechanisms are suitable for dealing with variable-sized inputs and working on the most relevant parts of inputs. If the attention mechanism is executed to compute a single sequence representation, it will be called self-attention. Self-attention architecture is parallelizable across node-neighbor pairs and can be applied to tasks where the model needs to generalize to completely unseen graphs [49].

Despite GCNs in which all neighbors share fixed weights and contribute equally during information passing, graph attention networks assign different levels of importance to each neighborhood of a specific node. In graph attention networks, layers are stacked, and nodes can attend over their neighbors. Each node will hold a different weight in a neighborhood. The advantage of these networks is that they do not require any knowledge about the structure of the graph and any matrix operations [49]. Graph attention network (GAT) [49] uses multi-head attention to stabilize the learning process and boost performance by concatenating $n$ attention heads. However, using multi-head attention can have a large size of parameters. To address this issue, a relation-aware graph attention network (RAGAT) [56] is proposed, which defines relation-aware message passing functions parameterized by relation-specific network parameters and employs averaging instead of concatenating $n$ attention head. To validate the results of this model, the decoder (scoring function) uses two different decoders: ConvE and InteractE. R-GCN and RAGAT are able to transform each neighbor of an entity in terms of their relation. However, they struggle with over-parameterization issues, especially with a large number of relations. The over-parameterization problem is tried to be solved in [57] by eliminating the dedicated parameter introduction for specific relations. R-GAT, unlike classic GAT, considers relation features.

The other issue of GAT is they ignore relation features [18]. A novel embedding must incorporate relation and neighboring node features in the attention mechanism to solve this issue. KBGAT [18] is a multi-hop and semantically similar relation extraction in the knowledge graph's $n$-hop neighborhood of any given entity. At the same time that KBGAT concatenates entity and relation embeddings to calculate the attention values, r-GAT transforms them separately. Learning embeddings are performed through a linear transformation to obtain absolute attention value over the concatenation of entity and relation feature vectors corresponding to a particular triple. Although the results are impressive, the disadvantage of this method is the computational costs and requirement of pre-trained KGE as input for the neural network.

A novel aggregation of neighborhood strategy with a local structure for knowledge graph completion has been proposed by [58]. LSA-GAT uses local structures to derive a sophisticated representation that covers semantic and structural information. The combination of LSA-GAT, local structure representation module, feature fusion module, and CNN-based decoder showed significant results. However, aggregating neighborhood entities fails to effectively model the critical relationships and ignores the distinct aspects of entities and relations. DisenKGAT [59] tried to learn the disentangled representation of entities by using micro- and macro-disentanglement as property of the KG. The robustness of this model is it can work with different kinds of score functions.

HRAN [60] proposed an attention-based model for heterogeneous graph networks such as knowledge graphs which have various types of entities and relations by aggregating features from different semantic aspects and dedicating weights to the relation path. They aggregate the neighbor features of an entity first, and the importance of each relation-paths is learned through relation features. The extracted features are aggregated with learned weights and generate embedding representations. The node feature aggregation of this model is performed through graph convolution. Next, due to heterogeneity in KGs, an entity-level relation-path-based aggregation is used. In the relation-level aggregation step, a novel relation-based attention mechanism is proposed to obtain the importance of different relation paths.

DecentRL [43] is a KG representation learning approach that encodes each node from the embeddings of its neighbors. Unlike other GNNs which consider the representation of an entity itself and its neighbors, this approach can be generalized to represent unseen entities by just learning representations from their context neighbors. The idea of decentRL is based on averaging its neighbor embeddings, which decentralizes the semantic information of entities over their neighbors. DecentRL works based on a decentralized attention network (DAN). DAN and GAT have identical layers, but DAN has a decentralized structure. In this case, the entity participates in the attention scores and the aggregation of neighbors. If the entity is an open entity, then GAT generates the embedding completely random and will be almost meaningless.

In contrast, DAN generates the embedding of the entity without the requirement of its embedding. This shows DAN's robustness and more descriptive aspect than conventional GAT. This model is considered a prototype of a graph attention mechanism in an open-world setting. They proposed an efficient knowledge distillation algorithm for generating unseen entities. The results outperform entity alignment and entity prediction tasks compared to current models under open-world settings such as AlignE, GAT, and AliNet. Below the order of discussed models have been depicted in Figure 7.

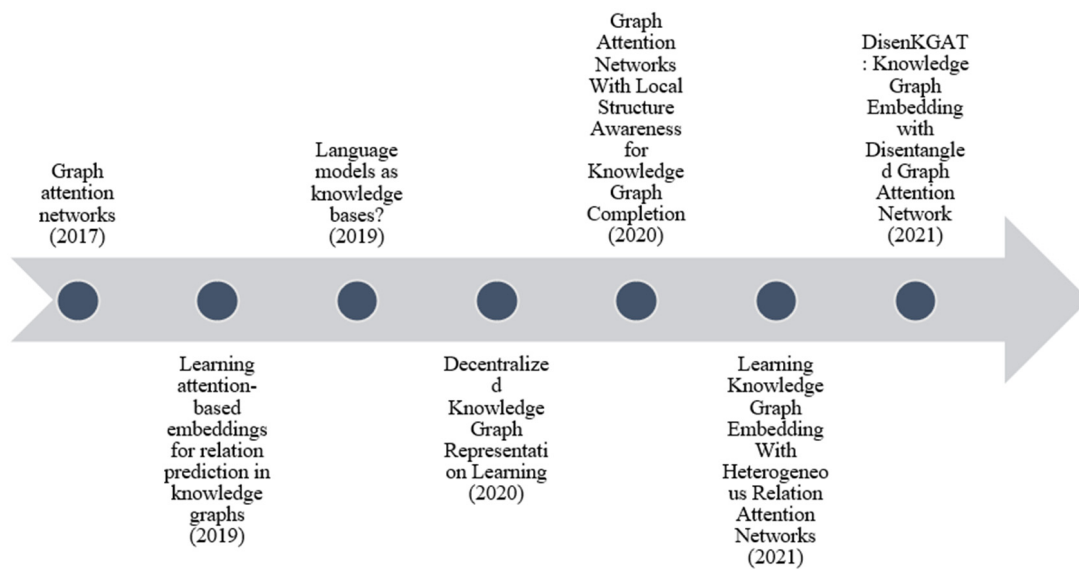

**Figure 7.** Attention neural network-based models' timeline.

Table 7 depicts all reviewed approaches which experimented with their models on the WN18RR and FB15k-237 datasets. The results show TransD and KGBERT have the best results on WN18RR and QuatE and DualE have better results on FB15k-237.

**Table 7.** Link prediction with different embedding settings. Lower values in MR and Higher Hits are better.

| Model | WN18RR | | FB15k-237 | |
|---|---|---|---|---|
| | MR | Hits@10 | MR | Hits@10 |
| TransE [13] | 3384 | 50.1 | 357 | 46.5 |
| TransH [61] | 2524 | 50.3 | 255 | 48.6 |
| TransR [61] | 3166 | 50.7 | 237 | 51.1 |
| TransD [61] | 276 | 50.7 | 246 | 48.4 |
| DistMult [61] | 3704 | 47.7 | 411 | 41.9 |
| ComplEx [61] | 3921 | 48.3 | 508 | 43.4 |
| Tucker [31] | - | 52.6 | - | 54.4 |
| ConvE [28] | 5277 | 48 | 246 | 49.1 |
| InteractE [37] | 5202 | 52.8 | 172 | 53.5 |
| ConvKB [39] | 3324 | 52.4 | 311 | 42.1 |

| Model | WN18RR | | FB15k-237 | |
|---|---|---|---|---|
| ConEx [38] | - | 55 | - | 55.5 |
| LSA-GAT [49] | 1947 | 44 | 273 | 60 |
| HARN [60] | 2113 | 54.2 | 156 | 54.1 |
| R-GCN [15] | - | - | - | 41.7 |
| RotatE-GCN [17] | - | 55.5 | - | 57.8 |
| TransE-GCN [17] | - | 47.7 | - | 50.8 |
| COMPGCN [16] | 3533 | 54.6 | 197 | 53.5 |
| RotatE [13] | 3384 | 50.1 | 177 | 53.3 |
| HAKE [23] | - | 58.2 | - | 54.2 |
| KG-BERT [61] | 97 | 52.4 | 153 | 42.0 |
| QuatE [13] | 2314 | 58.2 | 87 | 55 |
| DualE [30] | 2270 | 44.4 | 91 | 55.9 |
| DisenKGAT [59] | 1504 | 57.8 | 179 | 55.3 |
| RAGAT [56] | 2390 | 56.22 | 199 | 54.7 |
| KBGAT [18] | 1921 | 55.4 | 270 | 33.1 |
| Inverse Model [28] | 13,219 | 36 | 7148 | 1.2 |
| decentRL + TransE [43] | - | - | 159 | 52.1 |
| decentRL + DistMult [43] | - | - | 151 | 54.1 |
| RGCN + TransE [43] | - | - | 325 | 44.3 |
| RGCN + DistMult [43] | - | - | 230 | 49.9 |

*3.3. Pre-Trained Neural Network Models in Knowledge Graphs*

Knowledge graph construction is mainly supervised and requires humans to define all the facts manually, such as Wikidata or Freebase. Extracting the facts can also be performed with a semi-supervised approach, which still needs human supervision. With the evolution of language models such as BERT, outperforming results in various natural language tasks have been achieved. However, pre-trained language models (PLMs) struggle to capture rich knowledge. Existing PLMs learn helpful knowledge from unlabeled text and cannot capture the facts well because of the sparsity and complex forms in the text. In contrast, knowledge embedding models can represent relational facts in structured data rather than an unstructured text corpus. Recently some works have studied the applicability of pre-trained models in the context of KGs, which proved to be a promising solution.

The unified knowledge embedding and pre-trained language representation (KEPLER) [62] is proposed to integrate factual knowledge into the pre-trained language model and produce effective text-enhanced knowledge embedding. The textual entity descriptions are encoded with a pre-trained language model as their embeddings to optimize the KE and language modeling objectives jointly. In this approach, entities are encoded into vectors using their corresponding text. They produced Wikidata5M from Wikipedia data dump 2019 on two different settings, transductive and inductive. In a transductive setting, entities are shared, and triple sets are disjointed in train, test, and validation data. In contrast, in an inductive setting, the entities and triplets are mutually disjointed across train, validation, and test.

Petroni [63] presented an in-depth analysis of to what extent pre-trained language models can store factual and common-sense relational knowledge. The results on relational knowledge bases show that without fine-tuning, BERT provides competitive results compared to traditional methods in the case of relational knowledge. They also showed that BERT could outperform on open-domain question answering compared to supervised methods. Finally, they proved some specific types of factual knowledge are readily learned. However, this work evaluated knowledge to present and did not investigate the link prediction model in OKGs.

Wang et. al. [64] proposed an unsupervised end-to-end model named Match and Map (MAMA). It constructs KGs with a single forward passing of pre-trained language models. In the Match stage, a set of candidate facts from corpora will be created. The stored knowledge in the pre-trained language model will be matched with the target corpora at this stage and each of the extracted triples will be passed to the Map stage. In the Map stage, an OKG will be constructed with the candidate facts. If the constructed facts can be framed in a fixed KG schema, then they will be mapped according to the Wikidata schema. If the candidate is in an open schema, it will be partially mapped. This work's issue will be to increase the size of unmapped facts on a large scale and lead to performance problems.

Talukdar [65] showed BERT may not predict the correct entity for OKGs, but it can still predict type compatible entities well. The experiment result for entity linking was also the same [66]. As mentioned before, OKGs do not have an underlying ontology. Hence, providing type information is expensive and time-consuming, and BERT predictions can improve OKGs link prediction [65]. They applied BERT to improve OKG link prediction with a novel scoring function in this work. OKGIT aimed to use the unsupervised implicit type of information present in the pre-trained BERT model into OKG embeddings instead of explicit entity types present in an ontology. Results outperformed ConvE and CaRE.

Yao et. al. [61] proposed a knowledge graph using a BERT pre-trained model to improve the performance with rich language information by capturing rich semantic patterns from free text. BERT predicts whether two input sentences are consecutive or not. A sentence in original BERT can be an arbitrary span of contiguous text or word sequence [61]. The plausibility of a triple is calculated by considering the sentences of ($h$, $r$, $t$) as a single sequence. The model uses sentences of entities h and t to predict the relation r between them. This way, the knowledge graph completion task is converted into a sequence classification problem. The results show that in link prediction, it outperformed in comparison with TransE, TransR, TransH, TransD, and DistMult.

The critical issue of current approaches which use language models is that they are all trained based on available datasets like Wikipedia. As a result, if OKG wants to be specific in an area, there might be issues in accuracy, and it requires pre-train a language model related to that area.

## 4. Challenges in Knowledge Graphs

As of now, many challenges remain unsolved. These challenges include scalability, entity disambiguation, knowledge extraction from heterogeneous and unstructured data, and managing evolving knowledge management. Most current works assume static knowledge graphs and do not change the facts over time. A promising direction will be studying dynamic graph algorithms which can handle the addition of new edges over a continuous time. Most of the current representation learning lacks the capability of using multilingual KGs. The use of multi-source knowledge bases and multi-modality in knowledge graphs is still not well studied.

Other than the above-mentioned challenges, most current real-world knowledge graphs have low quality, and constructing a special domain knowledge graph is cumbersome.

Most currently available KG embeddings are applicable in ontological KGs rather than OKGs. Although most ontological KGs are canonicalized, OKGs are not. As an example, Donald Trump, President Trump, and Trump are the same. Recently some works have been proposed, but they are not still at the state-of-the-art point.

With the growth of graphs at scale, extraction of knowledge from multiple structured and unstructured sources is still challenging. Using supervised learning approaches needs human annotation, which is time-consuming. This leads the researchers to use unsupervised and semi-supervised methods.

Although some models have been proposed to provide additional semantic information, they still lack incorporating great semantic information. Additionally, the above-mentioned embedding models are not capable of knowledge inference. It seems graph neural network models are a promising approach that still needs more studies to improve the performance while reducing the complexity. However, jointly leveraging the power of both GCN models and knowledge graph completion methods for both entities and relations is still an open challenge.

## 5. Conclusions

KG provides an effective way of representing real-world relationships. KGE represents all the components of KG in vector form to represent the latent properties of the components. KGE methods as a key component of knowledge graphs started with translation-based models and continued by introducing more advanced approaches recently. The shortcomings of each approach have been expressed. This paper reviewed the main scoring functions which can result in different levels of expressiveness of facts. They are classified into translational, decompositional, neural network-based, and convolutional-based models. Unlike previous works which focused on the knowledge embedding models, we added how they could be integrated with graph neural networks to predict links. For this purpose, we introduced different graph neural network-based knowledge graph embedding. Overall, results show that the early models such as TransE although have faster computations, but they carry out less expressiveness in results.

On the other hand, using more complex scoring functions can also result in higher computational costs but better expressiveness. Graph neural network models mostly work on encoder-decoder based are one promising solution for link prediction and knowledge graph completion. Although many types of research have been performed, most of the current approaches are based on general datasets which are publicly available, and the results in special domains are not well studied or evaluated.

**Author Contributions:** Conceptualization, M.Z. and H.R. and M.R.; methodology, M.Z.; validation, M.Z., H.R. and M.R.; formal analysis, M.Z.; investigation, M.Z.; resources, M.Z.; writing—original draft preparation, M.Z.; writing—review and editing, H.R. and M.R.; visualization, M.Z.; supervision, H.R. and M.R. All authors have read and agreed to the published version of the manuscript.

**Funding:** This research received no external funding.

**Institutional Review Board Statement:** Not applicable.

**Informed Consent Statement:** Not applicable.

**Data Availability Statement:** The data included in this study are available upon request by contact with the first author.

**Conflicts of Interest:** The authors declare no conflict of interest.

## References

1. Stokman, F.N.; Vries, P.H.D. Structuring knowledge in a graph. In *Human-Computer Interaction*; Springer: Berlin/Heidelberg, Germany, 1988; pp. 186–206.
2. Noy, N.; Gao, Y.; Jain, A.; Narayanan, A.; Patterson, A.; Taylor, J. Industry-scale Knowledge Graphs: Lessons and Challenges: Five diverse technology companies show how it's done. *Queue* **2019**, *17*, 48–75.
3. Lenat, D.; Guha, R. Building large knowledge-based systems: Representation and inference in the CYC project. *Artif. Intell.* **1993**, *61*, 4152.
4. Auer, S.; Bizer, C.; Kobilarov, G.; Lehmann, J.; Cyganiak, R.; Ives, Z. Dbpedia: A nucleus for a web of open data. In *The Semantic Web*; Springer: Berlin/Heidelberg, Germany, 2007; pp. 722–735.
5. Bollacker, K.; Evans, C.; Paritosh, P.; Sturge, T.; Taylor, J. Freebase: A collaboratively created graph database for structuring human knowledge. In Proceedings of the 2008 ACM SIGMOD International Conference on Management of Data, Vancouver, BC, Canada, 10–12 June 2008.
6. Suchanek, F.M.; Kasneci, G.; Weikum. G. Yago: A core of semantic knowledge. In Proceedings of the 16th International Conference on World Wide Web, Banff, AB, Canada, 8–12 May 2007.
7. Vrandečić, D.; Krötzsch, M. Wikidata: A free collaborative knowledgebase. *Commun. ACM* **2014**, *57*, 78–85.
8. Wang, S.; Huang, C.; Li, J.; Yuan, Y.; Wang, F.-Y. Decentralized construction of knowledge graphs for deep recommender systems based on blockchain-powered smart contracts. *IEEE Access* **2019**, *7*, 136951–136961.

9. Antoine, B.; Usunier, N.; Garcia-Duran, A.; Weston, J.; Yakhnenko, O. Translating embeddings for modeling multi-relational data. In Proceedings of the 27th Annual Conference on Neural Information Processing Systems 2013, Lake Tahoe, NV, USA, 5–8 December 2013; Volume 26.
10. Wang, Z.; Zhang, J.; Feng, J.; Chen, Z. Knowledge graph embedding by translating on hyperplanes. In Proceedings of the AAAI Conference on Artificial Intelligence, Quebec City, QC, Canada, 27–31 July 2014.
11. Yang, B.; Yih, W.-t.; He, X.; Gao, J.; Deng, L. Embedding entities and relations for learning and inference in knowledge bases. *arXiv* **2014**, arXiv:1412.6575.
12. Trouillon, T.; Welbl, J.; Riedel, S.; Gaussier, É.; Bouchard, G. Complex embeddings for simple link prediction. In Proceedings of the International Conference on Machine Learning, New York, NY, USA, 19–24 June 2016.
13. Sun, Z.; Deng, Z.-H.; Nie, J.-Y.; Tang, J. Rotate: Knowledge graph embedding by relational rotation in complex space. *arXiv* **2019**, arXiv:1902.10197.
14. Yu, D.; Yang, Y.; Zhang, R.; Wu, Y. Knowledge embedding based graph convolutional network. In Proceedings of the Web Conference 2021, Ljubljana, Slovenia, 19–23 April 2021.
15. Schlichtkrull, M.; Kipf, T.N.; Bloem, P.; van den Berg, R.; Titov, I.; Welling, M. Modeling relational data with graph convolutional networks. In *European Semantic Web Conference*; Springer: Berlin/Heidelberg, Germany, 2018.
16. Vashishth, S.; Sanyal, S.; Nitin, V.; Talukdar, P. Composition-based multi-relational graph convolutional networks. *arXiv* **2019**, arXiv:1911.03082.
17. Cai, L.; Yan, B.; Mai, G.; Janowicz, K.; Zhu, R. TransGCN: Coupling transformation assumptions with graph convolutional networks for link prediction. In Proceedings of the 10th International Conference on Knowledge Capture, Marina Del Rey, CA, USA, 19–21 November 2019.
18. Nathani, D.; Chauhan, J.; Sharma, C.; Kaul, M. Learning attention-based embeddings for relation prediction in knowledge graphs. *arXiv* **2019**, arXiv:1906.01195.
19. Lin, Y.; Liu, Z.; Sun, M.; Liu, Y.; Zhu, X. Learning entity and relation embeddings for knowledge graph completion. In Proceedings of the Twenty-Ninth AAAI Conference on Artificial Intelligence, Austin, TX, USA, 25–30 January 2015.
20. Ji, G.; He, S.; Xu, L.; Liu, K.; Zhao, J. Knowledge graph embedding via dynamic mapping matrix. In Proceedings of the 53rd Annual Meeting of the Association for Computational Linguistics and the 7th International Joint Conference on Natural Language Processing, Beijing, China, 26–31 July 2015; Volume 1.
21. Fan, M.; Zhou, Q.; Chang, E.; Zheng, F. Transition-based knowledge graph embedding with relational mapping properties. In Proceedings of the 28th Pacific Asia Conference on Language, Information and Computing, Phuket, Thailand, 12–14 December 2014.
22. Ma, L.; Sun, P.; Lin, Z.; Wang, H. Composing knowledge graph embeddings via word embeddings. *arXiv* **2019**, arXiv:1909.03794.
23. Zhang, Z.; Cai, J.; Zhang, Y.; Wang, J. Learning hierarchy-aware knowledge graph embeddings for link prediction. In Proceedings of the AAAI Conference on Artificial Intelligence, New York, NY, USA, 7–12 February 2020.
24. Hogan, A.; Blomqvist, E.; Cochez, M.; d'Amato, C.; Melo, G.d.; Gutierrez, C.; Kirrane, S.; Gayo, J.E.L.; Navigli, R.; Neumaier, S. Knowledge graphs. In *Synthesis Lectures on Data, Semantics, and Knowledge*; Morgan & Claypool Publishers: San Rafael, CA, USA, 2021; Volume 12, pp. 1–257.
25. Nickel, M.; Tresp, V.; Kriegel. H.-P. A three-way model for collective learning on multi-relational data. In Proceedings of the 28th International Conference on Machine Learning, Bellevue, WA, USA, 28 June–2 July 2011.
26. Wang, Q.; Mao, Z.; Wang, B.; Guo, L. Knowledge graph embedding: A survey of approaches and applications. *IEEE Trans. Knowl. Data Eng.* **2017**, *29*, 2724–2743.
27. Bordes, A.; Glorot, X.; Weston, J.; Bengio, Y. A semantic matching energy function for learning with multi-relational data. *Mach. Learn.* **2014**, *94*, 233–259.
28. Dettmers, T.; Minervini, P.; Stenetorp, P.; Riedel, S. Convolutional 2d knowledge graph embeddings. In Proceedings of the AAAI Conference on Artificial Intelligence, New Orleans, LA, USA, 2–7 February 2018.
29. Zhang, S.; Tay, Y.; Yao, L.; Liu, Q. Quaternion knowledge graph embeddings. In Proceedings of the Annual Conference on Neural Information Processing Systems 2019, Vancouver, BC, Canada, 8–14 December 2019; Volume 32.
30. Cao, Z.; Xu, Q.; Yang, Z.; Cao, X.; Huang, Q. Dual quaternion knowledge graph embeddings. In Proceedings of the AAAI Conference on Artificial Intelligence, Online, 2–9 February 2021.
31. Balažević, I.; Allen, C.; Hospedales, T.M. Tucker: Tensor factorization for knowledge graph completion. *arXiv* **2019**, arXiv:1901.09590.
32. Wang, Q.; Wang, B.; Guo. L. Knowledge base completion using embeddings and rules. In Proceedings of the Twenty-Fourth International Joint Conference on Artificial Intelligence, Buenos Aires, Argentina, 25–31 July 2015.
33. Guo, S.; Wang, Q.; Wang, B.; Wang, L.; Guo, L. Semantically smooth knowledge graph embedding. In Proceedings of the 53rd Annual Meeting of the Association for Computational Linguistics and the 7th International Joint Conference on Natural Language Processing, Beijing, China, 26–31 July 2015; Volume 1.
34. Lin, Y.; Liu, Z.; Luan, H.; Sun, M.; Rao, S.; Liu, S. Modeling relation paths for representation learning of knowledge bases. *arXiv* **2015**, arXiv:1506.00379.

35. Socher, R.; Chen, D.; Manning, C.D.; Ng, A. Reasoning with neural tensor networks for knowledge base completion. In Proceedings of the 27th Annual Conference on Neural Information Processing Systems 2013, Lake Tahoe, NV, USA, 5–8 December 2013; Volume 26.
36. Balažević, I.; Allen, C.; Hospedales. T.M. Hypernetwork knowledge graph embeddings. In Proceedings of the International Conference on Artificial Neural Networks, Munich, Germany, 17–19 September 2019; Springer: Berlin/Heidelberg, Germany, 2019.
37. Vashishth, S.; Sanyal, S.; Nitin, V.; Agrawal, N.; Talukdar, P. Interacte: Improving convolution-based knowledge graph embeddings by increasing feature interactions. In Proceedings of the AAAI Conference on Artificial Intelligence, New York, NY, USA, 7–12 February 2020.
38. Yu, D.; Zhu, C.; Yang, Y.; Zeng, M. Jaket: Joint pre-training of knowledge graph and language understanding. *arXiv* **2020**, arXiv:2010.00796.
39. Nguyen, D.Q.; Nguyen, T.D.; Nguyen, D.Q.; Phung, D. A novel embedding model for knowledge base completion based on convolutional neural network. *arXiv* **2017**, arXiv:1712.02121.
40. Demir, C.; Ngomo. A.-C.N. Convolutional complex knowledge graph embeddings. In *European Semantic Web Conference*; Springer: Berlin/Heidelberg, Germany, 2021.
41. Song, T.; Luo, J.; Huang, L. Rot-pro: Modeling transitivity by projection in knowledge graph embedding. *Adv. Neural Inf. Process. Syst.* **2021**, *34*, 24695–24706.
42. Li, Y.; Tarlow, D.; Brockschmidt, M.; Zemel, R. Gated graph sequence neural networks. *arXiv* **2015**, arXiv:1511.05493.
43. Guo, L.; Wang, W.; Sun, Z.; Liu, C.; Hu, W. Decentralized Knowledge Graph Representation Learning. *arXiv* **2020**, arXiv:2010.08114.
44. Hogan, A.; Blomqvist, E.; Cochez, M.; D'amato, C.; Melo, G.D.; Gutierrez, C.; Kirrane, S.; Gayo, J.E.L.; Navigli, R.; Neumaier, S.; et al. Knowledge Graphs. *ACM Comput. Surv.* **2021**, *54*, 71.
45. Sperduti, A.; Starita, A. Supervised neural networks for the classification of structures. *IEEE Trans. Neural Netw.* **1997**, *8*, 714–735.
46. Gori, M.; Monfardini, G.; Scarselli. F. A new model for learning in graph domains. In Proceedings of the 2005 IEEE International Joint Conference on Neural Networks, Montreal, QC, Canada, 31 July–4 August 2005.
47. Hamilton, W.; Ying, Z.; Leskovec, J. Inductive representation learning on large graphs. In Proceedings of the Annual Conference on Neural Information Processing Systems 2017, Long Beach, CA, USA, 4–9 December 2017; Volume 30.
48. Abboud, R.; Ceylan, I.; Lukasiewicz, T.; Salvatori, T. Boxe: A box embedding model for knowledge base completion. *Adv. Neural Inf. Process. Syst.* **2020**, *33*, 9649–9661.
49. Veličković, P.; Cucurull, G.; Casanova, A.; Romero, A.; Lio, P.; Bengio, Y. Graph attention networks. *arXiv* **2017**, arXiv:1710.10903.
50. Marcheggiani, D.; Titov, I. Encoding sentences with graph convolutional networks for semantic role labeling. *arXiv* **2017**, arXiv:1703.04826.
51. Shang, C.; Tang, Y.; Huang, J.; Bi, J.; He, X.; Zhou, B. End-to-end structure-aware convolutional networks for knowledge base completion. In Proceedings of the AAAI Conference on Artificial Intelligence, Honolulu, HI, USA, 27 January–1 February 2019.
52. Li, J.; Shomer, H.; Ding, J.; Wang, Y.; Ma, Y.; Shah, N.; Tang, J.; Yin, D. Are Graph Neural Networks Really Helpful for Knowledge Graph Completion? *arXiv* **2022**, arXiv:2205.10652.
53. Kazemi, S.M.; Poole, D. Simple embedding for link prediction in knowledge graphs. In Proceedings of the Thirty-Second Annual Conference on Neural Information Processing Systems, Montréal, QC, Canada, 3–8 December 2018; Volume 31.
54. Tian, A.; Zhang, C.; Rang, M.; Yang, X.; Zhan, Z. RA-GCN: Relational aggregation graph convolutional network for knowledge graph completion. In Proceedings of the 2020 12th International Conference on Machine Learning and Computing, Shenzhen, China, 15–17 February 2020.
55. Ying, R.; He, R.; Chen, K.; Eksombatchai, P.; Hamilton, W.L.; Leskovec, J. Graph convolutional neural networks for web-scale recommender systems. In Proceedings of the 24th ACM SIGKDD International Conference on Knowledge Discovery & Data Mining, London, UK, 19–23 August 2018.
56. Liu, X.; Tan, H.; Chen, Q.; Lin, G. RAGAT: Relation aware graph attention network for knowledge graph completion. *IEEE Access* **2021**, *9*, 20840–20849.
57. Chen, M.; Zhang, Y.; Kou, X.; Li, Y.; Zhang, Y. r-GAT: Relational Graph Attention Network for Multi-Relational Graphs. *arXiv* **2021**, arXiv:2109.05922.
58. Ji, K.; Hui, B.; Luo, G. Graph attention networks with local structure awareness for knowledge graph completion. *IEEE Access* **2020**, *8*, 224860–224870.
59. Wu, J.; Shi, W.; Cao, X.; Chen, J.; Lei, W.; Zhang, F.; Wu, W.; He, X. DisenKGAT: Knowledge Graph Embedding with Disentangled Graph Attention Network. In Proceedings of the 30th ACM International Conference on Information & Knowledge Management, Gold Coast, Australia, 1–5 November 2021.
60. Li, Z.; Liu, H.; Zhang, Z.; Liu, T.; Xiong, N.N. Learning knowledge graph embedding with heterogeneous relation attention networks. *IEEE Trans. Neural Netw. Learn. Syst.* **2021**, *33*, 3961–3973.
61. Yao, L.; Mao, C.; Luo, Y. KG-BERT: BERT for knowledge graph completion. *arXiv* **2019**, arXiv:1909.03193.
62. Wang, X.; Gao, T.; Zhu, Z.; Zhang, Z.; Liu, Z.; Li, J.; Tang, J. KEPLER: A Unified Model for Knowledge Embedding and Pre-trained Language Representation. *Trans. Assoc. Comput. Linguist.* **2021**, *9*, 176–194.

63. Petroni, F.; Rocktäschel, T.; Lewis, P.; Bakhtin, A.; Wu, Y.; Miller, A.H.; Riedel, S. Language models as knowledge bases? *arXiv* **2019**, arXiv:1909.01066.
64. Wang, C.; Liu, X.; Song, D. Language models are open knowledge graphs. *arXiv* **2020**, arXiv:2010.11967.
65. Talukdar, P.P. OKGIT: Open Knowledge Graph Link Prediction with Implicit Types. *arXiv* **2021**, arXiv:2106.12806.
66. Chen, S.; Wang, J.; Jiang, F.; Lin, C.-Y. Improving entity linking by modeling latent entity type information. In Proceedings of the AAAI Conference on Artificial Intelligence, New York, NY, USA, 7–12 February 2020.